\documentclass[a4paper,11pt]{article}
\usepackage{pos}

\usepackage{subcaption}

\title{Forecasting Seismic Waveforms: A Deep Learning Approach for Einstein Telescope}

\author*[a]{Waleed Esmail}
\author[a]{Alexander Kappes}
\author[b]{Stuart Russell}
\author[b]{Christine Thomas}

\affiliation[a]{Institut f\"ur Kernphysik, Universit\"at M\"unster,\\
  Wilhelm-Klemm-Straße 9, 48149, M\"unster}

\affiliation[b]{Institut f\"ur Geophysik, Universit\"at M\"unster,\\
  Corrensstraße 24, 48149, M\"unster}

\emailAdd{wesmail@uni-muenster.de}

\abstract{
We introduce \textit{SeismoGPT}, a transformer-based model for forecasting three-component seismic waveforms in the context of future gravitational wave detectors like the Einstein Telescope. The model is trained in an autoregressive setting and can operate on both single-station and array-based inputs. By learning temporal and spatial dependencies directly from waveform data, SeismoGPT captures realistic ground motion patterns and provides accurate short-term forecasts. Our results show that the model performs well within the immediate prediction window and gradually degrades further ahead, as expected in autoregressive systems. This approach lays the groundwork for data-driven seismic forecasting that could support Newtonian noise mitigation and real-time observatory control.}

\ConferenceLogo{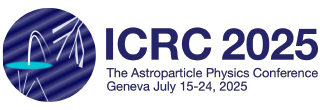}

\FullConference{39th International Cosmic Ray Conference (ICRC2025)\\
 15–24 July 2025\\
Geneva, Switzerland\\}

\begin{document}
\maketitle

\section{Introduction}
Seismic noise is a critical consideration for ground-based gravitational wave (GW) detectors, especially when it comes to limiting their sensitivity in the low-frequency band. These detectors aim to measure tiny variations in relative length, so any other source of motion, such as seismic vibrations, can introduce significant noise into the system.

Seismic noise refers to the persistent and variable ground vibrations that are present everywhere on Earth \cite{Trozzo:2022tar}. Seismic waves are classified into two main types: body waves (P and S waves that move through the interior of the Earth) and surface waves (Rayleigh and Love waves that travel along the surface and decay with depth).

In GW detectors, seismic noise couples into the system in two main ways:

\begin{itemize}
    \item Mechanical transmission through suspensions and isolation systems: Ground motion directly shakes the optical components. Even in underground detectors, seismic motion at 10 Hz can still be many orders of magnitude stronger than a GW signal \cite{Trozzo:2022tar}.
    \item Newtonian Noise (NN): This is subtler. Seismic waves change the density of the surrounding ground and air, which causes fluctuations in the gravitational field, which in turn, pull directly the detector’s test masses. \cite{Harms:2019dqi}. 
\end{itemize}

Given the significant impact of seismic noise, especially in the 0.1–10 Hz frequency band, ground-based GW detectors rely on highly sophisticated isolation systems to keep it under control \cite{Trozzo:2022tar}. These systems are designed to prevent ground vibrations from coupling into the test masses, and they come in two main forms: passive isolation, which uses mechanical filtering through pendulums and springs, and active isolation, which involves sensors and actuators to detect and cancel vibrations in real time. 

For future third-generation GW detectors, such as the Einstein Telescope (ET) \cite{Punturo:2010zz} and Cosmic Explorer (CE) \cite{Hall:2022dik}, the plan is to extend the detection band below 10 Hz, which will require significant upgrades to the suspension systems to improve seismic attenuation in the low-frequency band. 

In this article, we introduce a deep learning-based approach for predicting seismic waveforms. While the primary focus is on forecasting, this method has potential applications in seismic noise mitigation, especially for future GW detectors operating in the low-frequency band. We have the ET particularly in mind, but the approach is general and could be applied to other observatories as well. Such predictions could support active control systems, help with NN subtraction, or improve detector scheduling by anticipating periods of high seismic activity.

\section{Deep Learning-Based Seismic Waveform Prediction}
\subsection{Motivation and Background}
Traditional methods like Wiener filtering \cite{Koley:2024wev} have already been used in current detectors to subtract coherent seismic noise, especially in the context of NN mitigation. These filters rely on linear assumptions and require dense sensor arrays, and while they've shown promising results, they still have limitations. Their performance drops when the seismic field is non-stationary or when the correlations between sensors and the target change over time, which is often the case in real environments.

This is where deep learning can be a potentially significant enhancement. Instead of relying on fixed linear correlations, neural networks can learn non-linear patterns, handle complex temporal dynamics, and adapt better to changing conditions. The idea is not to replace existing systems, but to explore whether a data-driven approach can complement and enhance current noise mitigation strategies, especially in the low-frequency range where NN is expected to dominate. 

\subsection{Model Architecture}
\begin{figure}[ht]
    \centering
    \begin{subfigure}[b]{0.48\textwidth}
        \centering
        \includegraphics[scale=0.54]{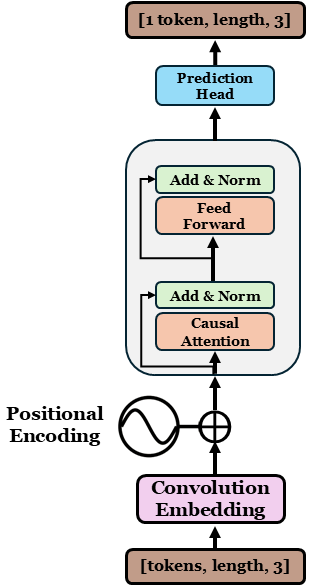}
        \caption{Single-station SeismoGPT model architecture.}
        \label{fig:seismogpt_single}
    \end{subfigure}
    \hfill
    \begin{subfigure}[b]{0.48\textwidth}
        \centering
        \includegraphics[width=\textwidth]{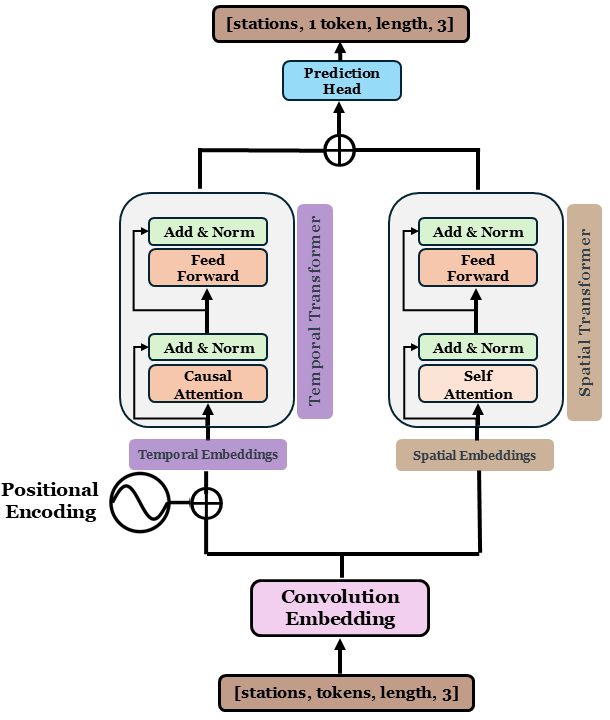}
        \caption{Array-based SeismoGPT model architecture.}
        \label{fig:seismogpt_array}
    \end{subfigure}
    \caption{Overview of the SeismoGPT architectures for seismic waveform forecasting. The single-station model (left) uses causal attention over temporal tokens. The array-based model (right) leverages both temporal and spatial attention using separate transformer branches.}
    \label{fig:seismogpt_models}
\end{figure}

We developed a transformer-based architecture, \textit{SeismoGPT}, shown in Figure \ref{fig:seismogpt_models}, for seismic waveform forecasting. The core idea is to frame the task as sequence modeling, where the model predicts the future evolution of a waveform given a short context window. We implement two versions of the model: one for single-station input and one for seismic arrays. Both share the same core architecture but differ in how spatial context is handled.

\textit{SeismoGPT} is based on the transformer encoder \cite{Vaswani:2017lxt}, which uses self-attention to capture long-range dependencies. In our autoregressive setup, we use \textit{causal attention}, where each token attends only to the current and past tokens, not the future.

Given an input sequence $X \in \mathbb{R}^{T \times d}$, causal self-attention is defined as:
\[
\text{Attention}(X) = \text{softmax} \left( \frac{Q K^\top}{\sqrt{d_k}} + M \right) V,
\]
where $Q = XW_Q$, $K = XW_K$, $V = XW_V$ are linear projections with learnable weights, and $M$ is a lower-triangular mask:
\[
M_{ij} =
\begin{cases}
0 & \text{if } j \leq i, \\
-\infty & \text{if } j > i.
\end{cases}
\]

This masking enforces the autoregressive property, ensuring predictions rely only on past and present context.

\subsubsection{Single-Station Model}
The single-station version of \textit{SeismoGPT} forecasts three-component seismic waveforms using a short context window from a single sensor. The input is a waveform tensor \( X \in \mathbb{R}^{T \times 3} \), where \( T \) is the number of time steps and the last dimension represents the Z, N, and E components. This waveform is split into \( N \) non-overlapping tokens of length \( L \), resulting in \( X_{\text{tok}} \in \mathbb{R}^{N \times L \times 3} \).

Each token is flattened along the temporal and channel dimensions and passed through a 1D convolutional embedding block, producing a high-dimensional sequence \( Z \in \mathbb{R}^{N \times d} \). To retain temporal structure, sinusoidal positional encoding \cite{Vaswani:2017lxt} is added to \( Z \), and the result is fed into a stack of transformer encoder layers with causal attention.

The encoded sequence \( H \in \mathbb{R}^{N \times d} \) is projected back to the original token shape using a linear prediction head:
\[
\hat{Y}_{\text{tok}} \in \mathbb{R}^{N \times L \times 3},
\]
which represents the predicted waveform for each token.

\subsubsection{Array-Based Model}
The array-based version of \textit{SeismoGPT} forecasts seismic waveforms by leveraging spatial correlations across a network of stations. It processes input data of shape \( X \in \mathbb{R}^{B \times S \times N \times L \times 3} \), where \( B \) is the batch size, \( S \) is the number of stations, \( N \) is the number of tokens per station, \( L \) is the token length, and 3 denotes the waveform components (Z, N, E).

Each token is flattened and embedded via a 2D convolutional network:

\[
X_{\text{flat}} \in \mathbb{R}^{B \times S \times N \times (L \cdot 3)} \rightarrow Z \in \mathbb{R}^{B \times S \times N \times d},
\]

where \( d \) is the embedding dimension. The embedded sequence is then reshaped into two branches: temporal and spatial. For temporal modeling, each station is treated independently:

\[
Z_{\text{temp}} \in \mathbb{R}^{(B \cdot S) \times N \times d},
\]

and for spatial modeling, each time step across all stations is grouped:

\[
Z_{\text{spat}} \in \mathbb{R}^{(B \cdot N) \times S \times d}.
\]

Each sequence is passed through its own positional encoding. The temporal branch uses causal self-attention, while the spatial branch applies full self-attention:

\[
H_{\text{temp}} = \text{Transformer}_{\text{temp}}(Z_{\text{temp}}), \quad
H_{\text{spat}} = \text{Transformer}_{\text{spat}}(Z_{\text{spat}}).
\]

The outputs are reshaped, summed, and projected back to waveform space using a linear head:

\[
\hat{Y}_{\text{tok}} \in \mathbb{R}^{B \times S \times (N \cdot L) \times 3},
\]

which represents the predicted 3-component waveform for each station over time.

\subsection{Training}

\subsubsection{Training Data}
To train and evaluate the proposed \textit{SeismoGPT} models, we generated a synthetic dataset of noise-free seismic waveforms using physically realistic source parameters and a global 1D Earth model. Each event is simulated using the Instaseis framework \cite{vanDriel2015} using the \texttt{ak135f\_2s} database from Syngine\footnote{\url{https://ds.iris.edu/ds/products/syngine/}}, which uses the ak135 1D velocity model \cite{Kennett1995}.

The source mechanisms are based on randomly generated moment tensors, created using a custom generator fitted to the statistical distribution of the Global Centroid Moment Tensor (GCMT) catalog\footnote{\url{https://www.globalcmt.org/}}. 


For each event, a set of synthetic three-component seismograms is simulated for receivers located at random azimuths and angular distances in the range 21--49 degrees. This range corresponds to teleseismic distances where waveform characteristics remain relatively stable across events. Limiting the source-receiver geometry to this controlled window helps reduce variability due to complex propagation effects and ensures more consistent training conditions for the model. The Instaseis interface returns velocity waveforms, which are then bandpass filtered according to the fundamental period of the database (2~s) and tapered to minimize edge effects. The simulation setup includes arrival time computation for $P$ and $S$ waves using the \texttt{TauPyModel}, ensuring consistency between waveform filtering and theoretical travel-time picking.

For the single-station setup, both the source and receiver locations are randomized, with the angular distance mentioned above. For the array-based model, we simulate a 16-station seismic network positioned near the proposed ET site in the Euregio Meuse-Rhine (EMR) region. The stations are arranged in a controlled geometric layout: 12 are distributed symmetrically around the vertices of an equilateral triangle (with a 10 km side length), and four are placed centrally. 

\subsubsection{Training Procedure}
To train the \textit{SeismoGPT} models, we employed an autoregressive forecasting setup where the model learns to predict the next token. The input waveform is tokenized into non-overlapping segments using a token length of 16 samples. For both models—the single-station and the array-based version, a fixed context window of 64 tokens (corresponding to 1024 samples) is used as input. Each token is passed through a convolutional embedding network followed by a transformer encoder consisting of 6 layers and 8 attention heads per layer.

During training, padding masks are randomly applied to the input token sequence, such that only a subset of the full context window is retained as input. This regularization technique encourages the model to generalize better by forcing it to perform predictions given variable context lengths. The target sequence is aligned accordingly, and the padding mask is used in the attention mechanism to ignore padded positions. The model is trained to minimize a simple mean squared error loss between the predicted and ground-truth waveform segments.

Both model versions are trained using the Adam optimizer \cite{Kingma:2014vow} with a learning rate of $5 \times 10^{-4}$. Learning rate scheduling is applied using a StepLR scheduler with a decay factor of 0.8 every 5 epochs. Training is performed for a maximum of 100 epochs with early stopping based on validation loss, using a patience of 3 epochs. The training runs are conducted on two NVIDIA RTX 4090 GPUs.

\section{Results}

\begin{figure}[t]
    \centering
    \includegraphics[scale=0.4]{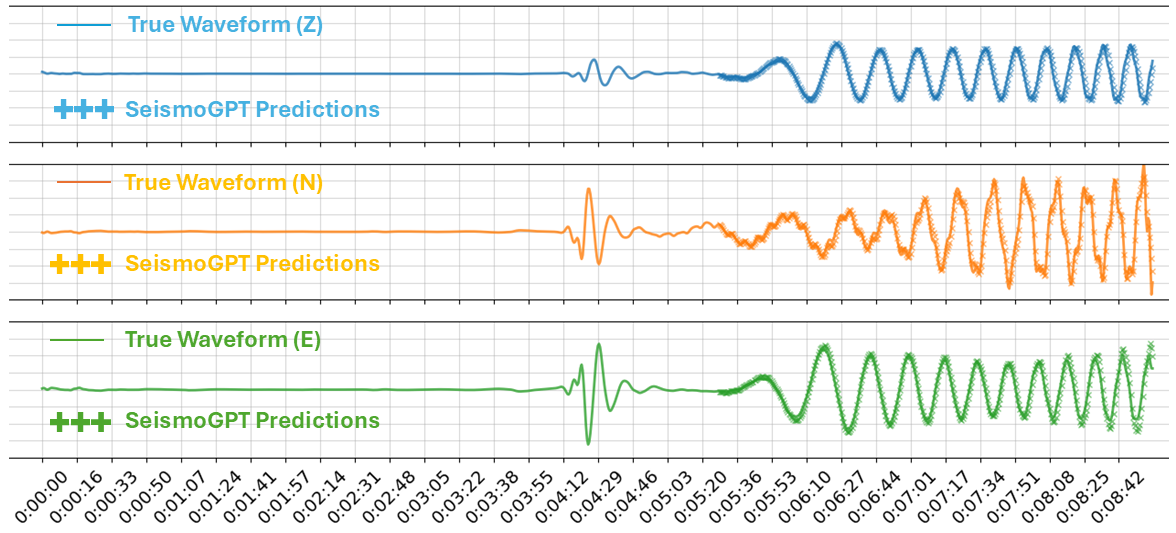}
    \caption{Prediction results from the single-station model. The input context consists of 40 tokens, and the model autoregressively predicts the remaining 24 tokens. The predicted waveform (crosses) is overlaid on the ground truth (solid lines) for each component (Z, N, E).}
    \label{fig:single_station}
\end{figure}

To assess the performance of \textit{SeismoGPT}, we evaluate both the single-station and array-based architectures. As a representative example, the model is provided with a context window of 40 tokens, each consisting of 16 samples, which corresponds to approximately 337.9 seconds of past waveform data (given a sampling rate of 1.9\,Hz). Based on this context, the model then autoregressively forecasts the next 24 tokens (384 samples, or approximately 202.1 seconds). This setup is used here as a showcase to demonstrate the model’s capabilities, but other context and prediction lengths are also possible.

\begin{figure}[t]
    \centering
    \includegraphics[width=\linewidth]{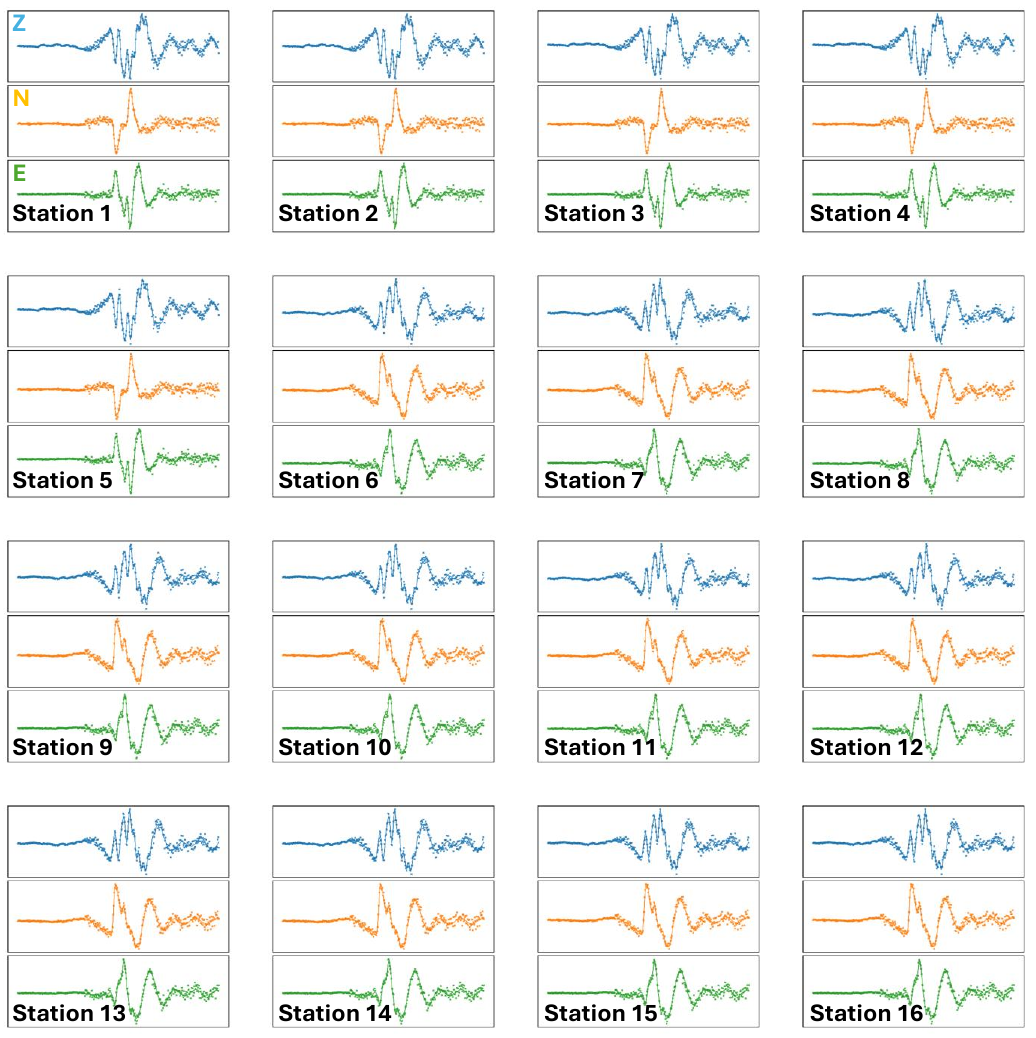}
    \caption{Prediction results from the array-based model for all 16 stations. Each station subplot shows the 3 components (Z, N, E) with the ground truth in solid lines and predictions as overlaid crosses. Same context and forecast settings as in Figure~\ref{fig:single_station}.}
    \label{fig:array_model}
\end{figure}

As illustrated in Figures ~\ref{fig:single_station} and ~\ref{fig:array_model}, the models demonstrate strong predictive accuracy during the initial forecast steps. This behavior is expected, as the model is still operating within a temporal range close to the input context. However, as prediction advances further into the future, the uncertainty increases and the accuracy gradually deteriorates. This is a direct consequence of the autoregressive decoding strategy: at each step, the model predicts a single token based on the previously observed. This predicted token is then appended to the context and used to forecast the next one. As a result, errors tend to accumulate over time, propagating from early inaccuracies and compounding with each subsequent step.

Despite this challenge, both models are able to capture the general waveform structure, including phase arrivals and oscillatory features, with reasonable fidelity. Notably, the array-based model benefits from the spatial redundancy across stations, which leads to more stable predictions compared to the single-station setting—especially in the later stages of the forecast window.

\section{Conclusion}
In this work, we introduced SeismoGPT, a transformer-based model designed to forecast seismic waveforms using both single-station and array-based input. The model operates autoregressively, using a fixed context window to generate future waveform segments token by token. While the core architecture remains the same in both settings, the array-based version takes advantage of spatial correlations across the seismic network to improve long-term stability. Our results show that the model performs well in predicting the general structure of the waveforms, particularly in the early stages of the forecast window. As expected, prediction quality degrades further into the future due to the autoregressive nature of the setup, where small errors can accumulate over time. That said, the overall performance is encouraging. The network is able to learn key waveform features and produce realistic forecasts even several minutes ahead. This opens the door to future applications where seismic forecasting could assist in active noise mitigation strategies, NN subtraction, or even real-time operational planning for gravitational wave observatories like the ET. 

\section*{Acknowledgments}
 This work is funded by the ErUM-WAVE project 05D2022, “ErUM-Wave: Antizipation 3-dimensionaler Wellenfelder”, which is supported by the German Federal Ministry of Education and Research (BMBF).

\end{document}